%% file: main.tex
\documentclass[11pt,a4paper]{article}
\usepackage[hyperref]{emnlp2018}
\usepackage{times}
\usepackage[normalem]{ulem} 
\usepackage{latexsym}
\usepackage{amsmath}
\usepackage{amssymb}

\usepackage{url}

\aclfinalcopy 
\def\aclpaperid{71} 

\usepackage{gnuplot-lua-tikz}

\def\hiden#1{}

\def\parcite#1{\cite{#1}} 
\def\perscite#1{\newcite{#1}} 
\def\inparcite#1{\citealp{#1}} 
\def\significantmark{$\ddagger$}
\def\significant{\rlap{$\ddagger$}} 

\def\task#1{\textsc{#1}}

\title{Trivial Transfer Learning for Low-Resource Neural Machine Translation}

\author{Tom Kocmi
        \qquad Ond{\v{r}}ej Bojar
		\\ \\
        Charles University, Faculty of Mathematics and Physics \\
        Institute of Formal and Applied Linguistics \\
        Malostransk{\'{e}} n{\'{a}}m{\v{e}}st{\'{\i}} 25, 118 00 Prague, Czech Republic \\
         {\tt <surname>@ufal.mff.cuni.cz}}

\date{}

\begin{document}
\maketitle
\begin{abstract}
Transfer learning has been proven as an effective technique for neural
machine translation under low-resource conditions. 
Existing methods require a common target language, language relatedness, 
or specific training tricks and regimes. 
We present a simple transfer learning method, where 
we first train a
``parent'' model for a
high-resource language pair
and then continue the training on a low-resource pair only by replacing the
training corpus.
This ``child'' model performs significantly better than the
baseline trained for
low-resource pair only. We are the first to show this for targeting different languages, and we observe the improvements even for unrelated languages with different alphabets.

\end{abstract}

\section{Introduction}

Neural machine translation (NMT) has made a big leap in performance and became
the unquestionable winning approach in the past 
few years \parcite{bahdanau2015neural,
sutskever2014sequence,sennrich-EtAl:2017:WMT,vaswani:2017}. The main reason
behind the success of NMT in realistic conditions was the ability to handle
large vocabulary \parcite{bpe} and to utilize large monolingual data
\parcite{sennrich-haddow-birch:2016:monolingual}.
However, NMT still struggles if
the parallel data is insufficient (e.g. fewer than 1M parallel sentences),
producing fluent output unrelated to the source and performing much worse than
phrase-based machine translation \parcite{koehn-knowles:2017:NMT}.

Many strategies have been used in MT in the past for employing resources from
additional languages, see e.g.
\perscite{wu-wang:2007:ACLMain}, \perscite{nakov2012improving},
\perscite{elkholy-EtAl:2013:Short}, or \perscite{hoang:bojar:pivoting:2016}.
For NMT, a particularly promising approach is transfer learning or
``domain adaptation'' where the ``domains'' are the different languages.

For example, \perscite{zoph-EtAl:2016:EMNLP2016} train a ``parent'' model in a high-resource 
language pair, then use some of the trained weights as the initialization for a ``child''
model and further train it on the 
low-resource language pair. In \perscite{zoph-EtAl:2016:EMNLP2016}, the parent and child pairs shared the target
language (English) and a number of modifications of the training process were needed to
achieve an improvement in translation from
Hansa, Turkish, and Uzbek into English with the help of French-English data.

\perscite{Nguyen2017} explore a related scenario where the parent language pair is also low-resource but it is 
related to the child language pair. They improved the previous approach by
using a shared vocabulary of subword units (BPE, \inparcite{bpe}). Additionally,
they used transliteration to improve their results.

In this paper, we contribute empirical evidence that transfer learning for NMT can be simplified even further.
We leave out the restriction on relatedness of the languages
and extend the 
experiments to parent--child pairs where the target language changes. Moreover, we do not utilize any special modifications to
the training regime or data pre-preprocessing.

In contrast to previous work, we test the method with the Transformer model \parcite{vaswani:2017},
instead of the recurrent approaches \parcite{bahdanau2015neural}. As documented in 
e.g. \perscite{popel:bojar:transformer-training:2018} and anticipated in 
WMT18,\footnote{\url{http://www.statmt.org/wmt18/translation-task.html}} the Transformer model seems 
superior to other NMT approaches.

\section{Method Description}

The proposed method is extremely simple: We train the parent language pair for a number of iterations and switch 
the training corpus to the child language pair for the rest of the training, without resetting any of the training 
(hyper)parameters.

As such, this method is similar to the transfer learning proposed by
\perscite{zoph-EtAl:2016:EMNLP2016} but 
uses the shared vocabulary as in
\perscite{Nguyen2017}. The novelty is that we are removing the restriction about relatedness of the
language pairs, and in contrast to the previous papers, we show that this simple style of
transfer learning can be used on both sides (i.e. either the source or the target language), not only with the 
target language common to both parent and child model. In fact, the method is
effective also for fully unrelated language pairs.

Our method does not need any modification of existing  NMT frameworks. The only 
requirement is to use a shared vocabulary of subword units (we use wordpieces, \inparcite{zeroshop_TACL1081}) 
across both language pairs. This is achieved by learning wordpiece segmentation from the 
concatenated source and target sides of both the parent and child language pairs.
All other parameters of the model stay the same as for the standard NMT training.

During the training we first train the NMT model for the high-resource language
pair until convergence. This model is called ``parent''. After that, we train
the child model without any restart, i.e. only by changing the training corpora
to the low-resource language pair.

\subsection{Details on Shared Vocabulary}
\label{sec:shared_vocabulary}

Current NMT systems use vocabularies of subword units instead of whole words.
Using
subword units gives a balance between the flexibility of separate characters and
efficiency of whole words. It solves the out-of-vocabulary words problem and
reduces the vocabulary size. The majority of NMT systems use either
the byte pair encoding \parcite{bpe} or wordpieces \parcite{wu2016google}.
Given a training corpus and the desired maximal vocabulary size, either method
produces deterministic rules for word segmentation to achieve the fewest
possible splits.

Our method requires the vocabulary shared across both the parent (translating from
language XX to YY) and the child model (translating from AA to BB). This is obtained by
concatenating both training corpora into one corpus of sentences in
languages AA, BB, XX and YY.
\footnote{Having separate vocabularies for the parent and child and
switching from the XX-YY to AA-BB vocabulary when we switch the
training corpus leads on an expected drop in performance. Independent
vocabularies use different IDs even for identical subwords and the network
cannot rely on any of its weights from the parent training.
}

Due to our focus on low-resource language pairs, we decided to generate
the vocabulary in a balanced way by selecting the same amount of sentences from both language
pairs. We thus use the same number of sentence pairs of the parent corpus as
there are in the child corpus.

We did not experiment with any other balancing of the vocabulary. Future
research 
could also investigate the
impact of using only the child corpus for vocabulary generation or various amounts
of used sentences.

We generated vocabularies aiming at 32k subword types. The exact size of the
vocabulary varies from 26.1k to 34.8k. All
experiments of a given language set use the same vocabulary.
Vocabulary overlap in each language set is further studied in Section \ref{sec:vocab_analysis}.

\section{Model Description}

We use the Transformer sequence-to-sequence model \parcite{vaswani:2017} as implemented in 
Tensor2Tensor \parcite{tensor2tensor} version 1.4.2. Our models are based on the 
``big single GPU'' configuration as defined in the paper. To fit the model to
our GPUs (NVIDIA GeForce GTX 1080 Ti with 11 GB RAM), we set the
batch size to 2300 tokens and limit sentence length to 100 wordpieces. 

We use exponential learning rate decay with the starting learning rate of 0.2 and 32000 warm up steps and Adam optimized. In our experiments, we find that it is undesirable to reset learning rate as it leads to the loss of the performance from the parent model. Therefore the transfer learning is handled only by changing the training corpora and nothing else.

Decoding uses the beam size of 8 and the length normalization penalty is set to 1.

The models were trained for 1M steps (approx. 140 hours), which was sufficient
for models to converge to the best performance. We selected the model with the best
performance on the development test for the final evaluation on the testset.

\section{Datasets}

In our experiments, we compare low-resource and high-resource language pairs
spanning two orders of magnitude of training data sizes. We consider Estonian (\task{ET})
and Slovak (\task{SK}) as low-resource languages compared to the Finnish (\task{FI}) and Czech (\task{CS})
counterparts. 

The choice of languages was closely related to the languages in this year's WMT
2018 shared tasks. In particular, Estonian and Finnish (paired with English)
were suggested as the main focus for their relatedness. We added Czech and Slovak
as another closely related language pair. Russian (\task{RU}) for the parent model was
chosen for two reasons: (1) written in Cyrillic, there will be hardly any
intersection in the shared vocabulary with the child language pairs, and (2)
previous work uses 
transliteration to handle
Russian, which is a nice contrast to our work. Finally, we added Arabic (\task{AR}), French (\task{FR})
and Spanish (\task{ES}) for experiments with unrelated languages.

The sizes of the training datasets are in Table \ref{table:dataset_sizes}.

\begin{table}[t]
\begin{center}
\small
\begin{tabular}{lr|rr|rr} 
Lang. & Sent.  & \multicolumn{2}{c}{Words} & \multicolumn{2}{c}{Vocabulary} \\
pair  & pairs  &  First & Second & First & Second  \\
\hline
ET,EN & 0.8 M & 14 M & 20 M & 631 k & 220 k \\
FI,EN & 2.8 M & 44 M  & 64 M & 1697 k & 545 k \\
SK,EN & 4.3 M & 82 M & 95 M & 1059 k & 610 k \\
RU,EN & 12.6 M & 297 M & 321 M & 2202 k & 3161 k \\
CS,EN & 40.1 M & 491 M & 563 M & 6253 k & 4130 k \\
\hline
AR,RU & 10.2 M & 243 M & 252 M & 2299 k & 2099 k \\
FR,RU & 10.0 M & 295 M & 238 M & 1339 k & 2045 k \\
ES,FR & 10.0 M & 297 M & 288 M & 1426 k & 1323 k \\
ES,RU & 10.0 M & 300 M & 235 M & 1433 k & 2032 k  \\
\end{tabular}
\end{center}
\caption{Datasets sizes overview. We consider Estonian and Slovak low-resource
languages in our paper. Word counts and vocabulary sizes are from the original
corpus, tokenizing only at whitespace and preserving the case.
}
\label{table:dataset_sizes}
\end{table}

If not specified otherwise we use training, development and test sets from 
WMT.\footnote{\url{http://www.statmt.org/wmt18/}}
Pairs with training sentences with
less than 
4 words or more than 75 words on either the source or the target side are removed to allow for a speedup of Transformer
by capping the maximal length and allowing a bigger batch size. The reduction of training
data is small and based on our experiments, it does not change the performance of the translation model.

We use the Europarl and Rapid corpora for Estonian-English. 
We disregard Paracrawl due to its noisiness. 
The development and test sets are from WMT news 2018.

The Finnish-English was prepared as in \perscite{ostling2017helsinki}, removing
Wikipedia headlines. The dev and test sets are from WMT news 2015.

For English-Czech, we use
all paralel data allowed in WMT2018 except Paracrawl. The main resource is
CzEng 1.7 
(the filtered version, \inparcite{bojar2016czeng}). The devset is WMT
newstest2011 and the testset is WMT newstest2017.

Slovak-English uses corpora from \perscite{galuvscakova2012improving}, detokenized by 
Moses.\footnote{\url{https://github.com/moses-smt/mosesdecoder}} WMT newstest2011
serves as the devset and testset.

The Russian-English training set was created from News Commentary, Yandex and UN
Corpus. As the devset, we use WMT newstest 2012.

The language pairs Arabic-Russian, French-Russian, Spanish-French and
Spanish-Russian were selected from UN corpus \parcite{ziemski2016united}, which
provides
over 10 million multi-parallel sentences in 6 languages.


\section{Results}

In this section, we present results of our approach.
Statistical significance of the winner (marked with \significantmark{}) is tested by paired bootstrap resampling against the
baseline (child-only) setup (1000 samples,
conf. level 0.05; \inparcite{bootstrap-koehn:2004}).

As customary, we label the models with the pair of the source and target
language codes,
for example the English-to-Estonian translation model is denoted by \task{ENET}.

The vocabularies are generated as described in \ref{sec:shared_vocabulary}
separately for each experimented combination of parent and child. The same
vocabulary is used whenever the parent and child use the same set of languages,
i.e.
disregarding the translation direction and model stage (parent or child).

\subsection{English as the Common Language}
\label{sec:english-as-the-common}

Table \ref{tab:highresourceparent} summarizes our results for various combinations of
high-resource parent and low-resource child language pairs when English
is shared between the child and parent either in the encoder or in the decoder. 

We confirm that sharing the target language improves performance as previously shown
\cite{zoph-EtAl:2016:EMNLP2016,Nguyen2017}.
This gains up to 
2.44 BLEU absolute for ETEN  with the FIEN parent. 
Using only the parent (FIEN) model to translate the child (ETEN) test set
gives a miserable performance, confirming the need for transfer learning or
``finetuning''.

\begin{table}[t]
\begin{center}
\small
\begin{tabular}{l|c|cc}
               &           & \multicolumn{2}{c}{Baselines: Only} \\
Parent - Child  & Transfer & Child & Parent\\
\hline
enFI - enET & 19.74\significant{} & 17.03 & 2.32\\
FIen - ETen &  24.18\significant{} & 21.74 & 2.44\\
\textbf{enCS - enET} &  20.41\significant{} & 17.03 & 1.42\\
\textbf{enRU - enET} &  20.09\significant{} & 17.03 & 0.57\\
\textbf{RUen - ETen} &  23.54\significant{} & 21.74 & 0.80\\
enCS - enSK & 17.75\significant{} & 16.13 & 6.51\\
CSen - SKen & 22.42\significant{} & 19.19 & 11.62\\
\hline
enET - enFI & 20.07\significant{} & 19.50 & 1.81\\
ETen - FIen & 23.95 & 24.40 & 1.78\\
enSK - enCS & 22.99 & 23.48\significant{} & 6.10\\
SKen - CSen & 28.20 & 29.61\significant{}  & 4.16\\
\end{tabular}
\end{center}
\caption{
Transfer learning with English reused either in source (encoder) or target
(decoder).
The column ``Transfer'' is our method, baselines correspond to training on one
of the corpora only.
Scores (BLEU) are always for the
child language pair and they are comparable only within lines or when the child
language pair is the same. ``Unrelated'' language pairs in bold. Upper part:
parent larger, lower part: child larger. (``EN'' lowercased
just to stand out.)}
\label{tab:highresourceparent}
\end{table}

A novel result is that the method works also for sharing the source
language, improving ENET by up to 2.71 BLEU thanks to ENFI parent.

Furthermore, the improvement is not restricted only to related languages as
Estonian and
Finnish as shown in previous works. Unrelated language pairs (shown in bold in
Table \ref{tab:highresourceparent}) like Czech and Estonian work too and in some cases 
even better than with the related datasets. We reach an
improvement of 3.38 BLEU for ENET when parent model was ENCS, compared to improvement
of 2.71 from ENFI parent. This statistically significant improvement contradicts 
\perscite{dabre2017empirical} who concluded that the more related the
languages are, the better transfer learning works. 
We see it as an indication that the size of the parent training set is more important than relatedness of languages.

The results with Russian parent for Estonian child (both directions) show that
transliteration is also
not necessary. Because there is no vocabulary sharing between Russian Cyrilic and
Estonian Latin (except numbers and punctuation, see
Section \ref{sec:vocab_analysis} for
further details), the improvement
could be attributed to a better coverage of English; an effect 
similar to domain adaptation.

On the other hand, this transfer learning works well only when the parent has more
training data than the child. As presented in the bottom part of
Table \ref{tab:highresourceparent}, low-resource parents do not generally improve the
performance of better-resourced childs and sometimes, they even (significantly) decrease it.
This is another indication, that the most important is the size of the parent
corpus compared to the child one.

The baselines are either models trained purely on the child parallel data or
only on the parent data. The second baseline only indicates the
relatedness of languages because it is only tested but never trained on the child
language pair. Also, we do not add any language tag as in
\perscite{zeroshop_TACL1081}. This
also highlights that the improvement of our method cannot be directly attributed
to the relatedness of languages: e.g. Czech and Slovak are much more similar
than Czech and Estonian (Parent Only BLEU of translation out of English is
6.51 compared to 1.42) and yet the gain from transfer learning is larger
for Estonian (+3.38) than from Slovak (+1.62).

\begin{table}[t]
\begin{center}
\small
\begin{tabular}{c|cc}
Child Training Sents & Transfer BLEU & Baseline BLEU \\
\hline
800k & 19.74 & 17.03 \\
400k & 19.04 & 14.94 \\
200k & 17.95 & 11.96 \\
100k & 17.61 & 9.39\\
50k  & 15.95 & 5.74  \\
10k  & 12.46 & 1.95 \\
\end{tabular}
\end{center}
\caption{Maximal score reached by ENET child for decreasing sizes of child
training data, trained off an ENFI parent (all ENFI data are used and models are trained for 800k steps). The baselines use only the reduced ENET data.
}
\label{tab:simulated_lowresource}
\end{table}

\subsection{Simulated Very Low Resources}

In Table \ref{tab:simulated_lowresource}, we simulate very low-resource settings by
downscaling the data for the child model.
It is a common knowledge, that gains from transfer learning are more pronounced
for smaller childs. The point of Table \ref{tab:simulated_lowresource} is to
illustrate that our approach is applicable even to extremely small child setups,
with as few as 10k sentence pairs.
Our transfer learning (``start with a model for whatever parent pair'') may
thus resolve the issue
of applicability of NMT for low resource languages as pointed out by
\perscite{koehn-knowles:2017:NMT}.

\subsection{Parent Convergence}

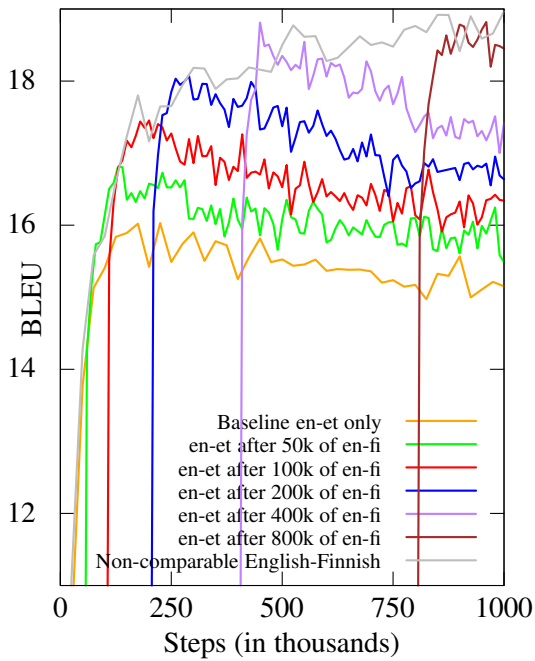
\begin{figure}
\begin{center}
\input{progress.tex}
\end{center}
\caption{Learning curves on dev set for ENFI parent and ENET child
where the child model started training after various numbers of the parent's
training steps.}
\label{fig:progress}
\end{figure}

Figure \ref{fig:progress} compares the performance of the child model when trained
from various training stages of the parent model. The performance
of the child clearly correlates with the performance of the parent. Therefore,
it
is better to use a parent model that already converged and reached its best performance.

\subsection{Direction Swap in Parent and Child}

\begin{table}[t]
\begin{center}
\small
\begin{tabular}{l|cc||c}
Parent - Child &  Transfer & Baseline & Aligned\\
\hline
enFI - ETen & 22.75\significant{} & 21.74 & 24.18\\
FIen - enET & 18.19\significant{} & 17.03 & 19.74\\
enRU - ETen & 23.12\significant{} & 21.74 & 23.54 \\
enCS - ETen & 22.80\significant{} & 21.74 & not run \\
RUen - enET & 18.16\significant{} & 17.03 & 20.09\\
\hline
enET - ETen & 22.04\significant{} & 21.74 & 21.74\\
ETen - enET & 17.46 & 17.03 & 17.03\\
\end{tabular}
\end{center}
\caption{Results of child following a parent with swapped direction.
``Baseline'' is child-only training. ``Aligned'' is the more natural setup with
English appearing on the ``correct'' side of the parent, the numbers in this
column thus correspond to those in Table \ref{tab:highresourceparent}. 
}
\label{tab:shared_english}
\end{table}

Relaxing the setup in Section \ref{sec:english-as-the-common}, we now allow a mismatch in translation direction of
the parent and child. The parent XX-EN is thus followed by an EN-YY child or
vice versa.
It is important to note that Transformer shares word embeddings for the source
and target side. The gain can be thus due to better English word
embeddings, but definitely not due to a better English language model. It would
be interesting to study the effect of not sharing the embeddings but we leave it
for some future work.

The results in Table \ref{tab:shared_english} document that
an improvement can be reached even when none of the involved languages is reused
on the same side. This interesting result should be studied in more
detail. \perscite{firat-cho-bengio:etal:2016} hinted possible gains
even when both languages are distinct from the low-resource
languages but in a multilingual setting.
Not surprisingly, the improvements are better when the common language
is aligned.

The bottom part of Table \ref{tab:shared_english} shows a particularly interesting
trick: the parent is not any high-resource pair but the very same EN-ET corpus
with source and target swapped. We see gains in both directions, although
not always statistically significant.
Future work should investigate if this 
performance boost is possible even for high-resource languages. Similar behavior
has been shown in \perscite{niu-denkowski-carpuat:2018:WNMT2018}, where in
contrast to our work they mixed the data together and added an artificial token indicating the target language.

\subsection{No Language in Common}

\begin{table}[t]
\begin{center}
\small
\begin{tabular}{l|cc}
Parent - Child &  Transfer & Baseline\\
\hline
ARRU - ETEN & 22.23 & 21.74 \\
ESFR - ETEN & 22.24\significant{} & 21.74 \\
ESRU - ETEN & 22.52\significant{} & 21.74 \\
FRRU - ETEN & 22.40\significant{} & 21.74 \\
\end{tabular}
\end{center}
\caption{Transfer learning with parent and child not sharing any language.}
\label{tab:no_shared}
\end{table}

Our final set of experiments examines the performance of ETEN child trained off
parents in totally unrelated language pairs. Without any common language,
the gains cannot be attributed, e.g., to the shared English word
embeddings. The vocabulary overlap is mostly due to short n-grams or numbers and punctuations.

We see gains from transfer learning in all cases, mostly significant. The only
non-significant gain is from Arabic-Russian
which does not share the
script with the child Latin at all. (Sharing of punctuation and numbers is
possible across all the tested scripts.) The gains are quite similar (+0.49--+0.78 BLEU), supporting our
assumption that the
main factor is the size of the parent (here, all have 10M sentence pairs) rather
than language relatedness.

\section{Analysis}

Here we provide a rather initial analysis of the sources of the gains.

\subsection{Vocabulary Overlap}
\label{sec:vocab_analysis}

Out method relies on the vocabulary estimated jointly from the child and parent
model. In Transformer, the vocabulary is even shared
across encoder and decoder. With a large overlap, we could expect a lot of
``information reuse'' between the parent and the child.

Since the subword vocabulary depends on the training corpora, a little clarification is
needed. We take the vocabulary of subword units as created e.g. for ENRU-ENET
experiments, see Section \ref{sec:shared_vocabulary}. This vocabulary
contains 28.2k subwords in total. We then process the training corpora for
each of the languages with this shared vocabulary, ignore all subwords that
appear less than 10 times in each of the languages (these subwords will have
little to no impact on the result of the training) and break down the total
28.2k subwords into classes depending on the languages in which
the particular subword was observed, see Table \ref{tab:enruet-breakdown}.

\def\yes{\checkmark}
\def\no{-}
\begin{table}[t]
\small
\begin{center}
\begin{tabular}{cccr}
ET & EN & RU & \% Subwords \\
\hline
\yes & \no & \no & 29.93\% \\
\no & \yes & \no & 20.69\% \\
\no & \no & \yes & 29.03\% \\
\yes & \yes & \no & 10.06\% \\
\no & \yes & \yes & 1.39\% \\
\yes & \no & \yes & 0.00\% \\
\yes & \yes & \yes & 8.89\% \\
\hline
\multicolumn{3}{l}{Total} & 28.2k (100\%) \\
\hline
\hline
\multicolumn{3}{l}{From parent} & 41.03\%
\end{tabular}
\end{center}
\caption{
Breakdown of subword vocabulary of experiments involving ET, EN and RU.}
\label{tab:enruet-breakdown}
\end{table}

We see that the vocabulary is reasonably balanced, with each language having
20--30\% of subwords unique to it. English and Estonian share 10\% subwords not
seen in Russian while Russian shares only 0--1.39\% of subwords with each of the
other languages. Overall 8.89\% of subwords are seen in all three languages.

A particularly interesting subset is the one where parent languages help the child
model, in other words subwords appearing anywhere in English and also tokens
common to Estonian and Russian. For this set of languages, this amounts to
20.69+10.06+1.39+0.0+8.89 = 41.03\%. We list this number on a separate line in
Table \ref{tab:enruet-breakdown}, ``From parent''. These subwords get their embeddings trained better
thanks to the parent model.

\begin{table}[t]
\begin{center}
\small
\begin{tabular}{l@{~~~}c@{~~~}r@{~~~}r}
Languages & Unique in a Lang. & In All & From Parent\\
\hline
ET-EN-FI & 24.4-18.2-26.2 & 19.5 & 49.4\\
ET-EN-RU & 29.9-20.7-29.0 & 8.9 &  41.0\\
ET-EN-CS & 29.6-17.5-21.2 & 20.3 & 49.2 \\
\hline
AR-RU-ET-EN & 28.6-27.7-21.2-9.1 & 4.6 & 6.2\\
ES-FR-ET-EN & 15.7-13.0-24.8-8.8 & 18.4 & 34.1\\
ES-RU-ET-EN & 14.7-31.1-21.3-9.3 & 6.0 & 21.4\\
FR-RU-ET-EN & 12.3-32.0-22.3-8.1 & 6.3 & 23.1\\
\end{tabular}
\end{center}
\caption{Summary of vocabulary overlaps for the various language sets. All
figures in \% of the shared vocabulary.}
\label{tab:vocab_stats}
\end{table}

Table \ref{tab:vocab_stats} summarizes this analysis for several language sets,
listing what portion of subwords is unique to individual languages in the set,
what portion is shared by all the languages and what portion of subwords
benefits from the parent training. We see a similar picture across the board,
only
AR-RU-ET-EN stands out with the very low number of subwords (6.2\%) available already in
the parent. The parent AR-RU thus offered very little word knowledge to the
child and yet lead to a gain in BLEU.

\subsection{Output Analysis}

Since we rely on automatic analysis, we need to prevent some potential
overestimations of translation quality due to BLEU.
For this, we took a closer
look at the baseline ENET model (BLEU of 17.03 in Table \ref{tab:highresourceparent}) and two ENET
childs derived from ENCS (BLEU of 20.41) and ENRU parent (BLEU 20.09).

Table \ref{tab:enet-details} confirms the improvements are not an artifact of uncased
BLEU. The gains are apparent with several (now cased) automatic
scores.

\begin{table}[t]
\footnotesize
\hspace*{-5mm}
\begin{tabular}{l@{}c@{~~}c@{~~}c@{~~}c@{~~}c@{~~}c}
           	& BLEU   	& nPER   	& nTER   	& nCDER  	& chrF3 	& nCharacTER \\
\hline
Base ENET  	& 16.13  	& 47.13  	& 32.45  	& 36.41  	& 48.38 	&      33.23 \\
ENRU+ENET  	& 19.10  	& 50.87  	& 36.10  	& 39.77  	& 52.12 	&      39.39 \\
ENCS+ENET  	& 19.30  	& 51.51  	& 36.84  	& 40.42  	& 52.71 	&      40.81
\end{tabular}
\caption{Various automatic scores on ENET test set. Scores prefixed ``n''
reported as $(1-\text{score})$ to make higher numbers better.}
\label{tab:enet-details}
\end{table}

As documented in Table \ref{tab:bleu-comps}, the improved outputs are considerably longer. In the table, we show also
individual $n$-gram precisions and brevity penalty (BP) of BLEU. The longer
output clearly helps to reduce the incurred BP but the improvements are also
apparent in $n$-gram precisions. In other words, the observed gain cannot be
attributed solely to producing longer outputs.

\begin{table}[t]
\footnotesize
\begin{tabular}{lcrr}
           	& Length	& BLEU Components     	& BP     \\
\hline               
Base ENET  	& 35326	& 48.1/21.3/11.3/6.4  	& 0.979   \\
ENRU+ENET  	& 35979	& 51.0/24.2/13.5/8.0  	& 0.998   \\
ENCS+ENET  	& 35921	& 51.7/24.6/13.7/8.1  	& 0.996  
\end{tabular}
\caption{Candidate total length, BLEU $n$-gram precisions and brevity penalty
(BP).
The reference length in the matching tokenization was 36062.}
\label{tab:bleu-comps}
\end{table}

\begin{table}
\begin{tabular}{lrr}
        	& ENRU+ENET         	& ENCS+ENET  	\\
\hline
rb      	& 15902 (44.2 \%)   	& 15924 (44.3 \%) \\
-       	& 9635 (26.8 \%)    	& 9485 (26.4 \%) \\
b       	& 7209 (20.0 \%)    	& 7034 (19.6 \%) \\
r       	& 3233 (9.0 \%)     	& 3478 (9.7 \%) \\
Total  	& 35979 (100.0 \%)  	& 35921 (100.0 \%)
\end{tabular}
\caption{Comparison of improved outputs vs. the baseline and reference.}
\label{tab:token-anots}
\end{table}

Table \ref{tab:token-anots} explains the gains in unigram precisions by checking which
tokens in the improved outputs (the parent followed by the child) were present
also in the baseline (child-only, denoted ``b'' in Table \ref{tab:token-anots}) and/or
confirmed by the reference (denoted ``r''). 
We see that about 44+20\% of tokens of improved outputs can be seen as
``unchanged" compared to the baseline because they appear already in the
baseline output (``b''). (The 44\% ``rb'' tokens are actually confirmed by the reference.)

The differing tokens are more interesting: ``-'' denotes the cases when the
improved system produced something different from the baseline and also from the
reference. Gains in BLEU are due to ``r'' tokens, i.e. tokens only in the
improved outputs and the reference but not the baseline ``b''. For both parent
setups, there are about
9--9.7 \% of such tokens. We looked at these 3.2k and 3.5k tokens and we have to
conclude that these are regular \emph{Estonian} words; no Czech or Russian leaks
to the output and the gains are \emph{not} due to simple token types common to
all the languages (punctuation, numbers or named entities). We see identical
BLEU gains even if we remove all such simple tokens from the candidates and
references. A better explanation of the gains thus still has to be sought for.

\section{Related Work}

\perscite{firat-cho-bengio:etal:2016} propose multi-way multi-lingual systems,
with the main goal of reducing the total number of parameters needed to cater
multiple source and target languages.
To keep all the language pairs ``active'' in the
model,
a special training schedule is needed. Otherwise, catastrophic forgetting would
remove the ability to translate among the languages trained earlier.

\perscite{zeroshop_TACL1081} is another multi-lingual approach: all translation pairs are simply used at once and the desired target language is indicated with a special 
token at the end of the source side. The model implicitly learns translation between many languages and it can even translate among language pairs never seen together.

Lack of parallel data can be tackled by unsupervised translation \parcite{artetxe2017unsupervised,lample2018phrase}. The general idea is to mix monolingual training of autoencoders for the source and target languages with translation trained on data translated by the previous iteration of the system.

When no parallel data are available, the trainset of closely related high-resource pair can be used with transliteration approach as described in \perscite{Karakanta2018}.

Aside from the common back-translation \parcite{sennrich-haddow-birch:2016:monolingual, kocmi-wmt-2018}, simple copying of target monolingual data back to source \parcite{currey2017copied} has been also shown to improve translation quality in low-data conditions.

Similar to transfer learning is also curriculum learning
\parcite{bengio2009curriculum, kocmi:ranlp}, where the training data are ordered
from foreign out-of-domain to the in-domain training examples.

\section{Conclusion}

We presented a simple method for transfer learning in neural machine translation
based on training a parent high-resource pair followed a low-resource language
pair dataset. The method works for shared source or target side as well as for
language pairs that do not share any of the translation sides. We observe gains
also from totally unrelated language pairs, although not always significant.

One interesting trick we propose for low-resource languages is to start training
in the opposite direction and swap to the main one afterwards.

The reasons for the gains are yet to be explained in detail but our observations
indicate that the key factor is the size of the parent corpus rather than e.g.
vocabulary overlaps.

\section*{Acknowledgments}

This study was supported in parts by the grants
 SVV~260~453,
 GAUK 8502/2016,
 and 18-24210S of the Czech Science Foundation.
This work has been using language resources and tools stored and distributed by the LINDAT/CLARIN project of the Ministry
of Education, Youth and Sports of the Czech
Republic (projects LM2015071 and OP VVV
VI CZ.02.1.01/0.0/0.0/16 013/0001781).

\bibliography{emnlp2018}
\bibliographystyle{acl_natbib_nourl}

\end{document}

%% file: progress.tex
\begin{tikzpicture}[gnuplot]
\path (0.000,0.000) rectangle (7.400,4.500);
\gpcolor{color=gp lt color border}
\gpsetlinetype{gp lt border}
\gpsetlinewidth{1.00}
\draw[gp path] (1.012,1.941)--(1.192,1.941);
\draw[gp path] (6.847,1.941)--(6.667,1.941);
\node[gp node right] at (0.828,1.941) { 12};
\draw[gp path] (1.012,3.852)--(1.192,3.852);
\draw[gp path] (6.847,3.852)--(6.667,3.852);
\node[gp node right] at (0.828,3.852) { 14};
\draw[gp path] (1.012,5.763)--(1.192,5.763);
\draw[gp path] (6.847,5.763)--(6.667,5.763);
\node[gp node right] at (0.828,5.763) { 16};
\draw[gp path] (1.012,7.674)--(1.192,7.674);
\draw[gp path] (6.847,7.674)--(6.667,7.674);
\node[gp node right] at (0.828,7.674) { 18};
\draw[gp path] (1.012,0.985)--(1.012,1.165);
\draw[gp path] (1.012,8.630)--(1.012,8.450);
\node[gp node center] at (1.012,0.677) { 0};
\draw[gp path] (2.471,0.985)--(2.471,1.165);
\draw[gp path] (2.471,8.630)--(2.471,8.450);
\node[gp node center] at (2.471,0.677) { 250};
\draw[gp path] (3.930,0.985)--(3.930,1.165);
\draw[gp path] (3.930,8.630)--(3.930,8.450);
\node[gp node center] at (3.930,0.677) { 500};
\draw[gp path] (5.388,0.985)--(5.388,1.165);
\draw[gp path] (5.388,8.630)--(5.388,8.450);
\node[gp node center] at (5.388,0.677) { 750};
\draw[gp path] (6.847,0.985)--(6.847,1.165);
\draw[gp path] (6.847,8.630)--(6.847,8.450);
\node[gp node center] at (6.847,0.677) { 1000};
\draw[gp path] (1.012,8.630)--(1.012,0.985)--(6.847,0.985)--(6.847,8.630)--cycle;
\node[gp node center,rotate=-270] at (0.674,4.807) {BLEU};
\node[gp node center] at (3.929,0.215) {Steps (in thousands)};
\node[gp node right,font={\fontsize{8pt}{9.6pt}\selectfont}] at (5.379,3.167) {Baseline en-et only};
\gpcolor{rgb color={1.000,0.647,0.000}}
\gpsetlinetype{gp lt plot 0}
\gpsetlinewidth{2.00}
\draw[gp path] (5.563,3.167)--(6.479,3.167);
\draw[gp path] (1.186,0.985)--(1.304,3.624)--(1.450,4.932)--(1.596,5.198)--(1.741,5.608)%
  --(1.887,5.660)--(2.033,5.783)--(2.179,5.211)--(2.325,5.791)--(2.471,5.275)--(2.617,5.513)%
  --(2.763,5.669)--(2.908,5.290)--(3.054,5.551)--(3.200,5.496)--(3.346,5.047)--(3.492,5.336)%
  --(3.638,5.587)--(3.784,5.273)--(3.930,5.309)--(4.075,5.225)--(4.221,5.242)--(4.367,5.307)%
  --(4.513,5.158)--(4.659,5.181)--(4.805,5.172)--(4.951,5.175)--(5.097,5.151)--(5.242,5.001)%
  --(5.388,5.034)--(5.534,4.948)--(5.680,4.971)--(5.826,4.782)--(5.972,5.119)--(6.118,5.091)%
  --(6.264,5.349)--(6.409,4.805)--(6.555,4.910)--(6.701,5.012)--(6.847,4.952);
\gpcolor{color=gp lt color border}
\node[gp node right,font={\fontsize{8pt}{9.6pt}\selectfont}] at (5.379,2.859) {en-et after 50k of en-fi};
\gpcolor{rgb color={0.000,1.000,0.000}}
\gpsetlinetype{gp lt plot 1}
\draw[gp path] (5.563,2.859)--(6.479,2.859);
\draw[gp path] (1.346,0.985)--(1.362,4.098)--(1.420,4.949)--(1.479,5.512)--(1.537,5.535)%
  --(1.596,5.896)--(1.654,6.246)--(1.712,6.322)--(1.771,6.539)--(1.829,6.532)--(1.887,6.100)%
  --(1.946,6.189)--(2.004,6.070)--(2.062,6.214)--(2.121,6.117)--(2.179,6.314)--(2.237,6.297)%
  --(2.296,6.255)--(2.354,6.459)--(2.412,6.265)--(2.471,6.264)--(2.529,6.359)--(2.587,6.264)%
  --(2.646,6.359)--(2.704,5.850)--(2.763,5.846)--(2.821,5.994)--(2.879,5.976)--(2.938,5.820)%
  --(2.996,6.025)--(3.054,5.778)--(3.113,5.977)--(3.171,5.614)--(3.229,5.798)--(3.288,6.117)%
  --(3.346,5.835)--(3.404,5.715)--(3.463,6.130)--(3.521,5.944)--(3.579,5.983)--(3.638,5.629)%
  --(3.696,5.989)--(3.754,5.932)--(3.813,5.781)--(3.871,5.437)--(3.930,5.839)--(3.988,6.100)%
  --(4.046,5.869)--(4.105,5.839)--(4.163,5.660)--(4.221,5.651)--(4.280,5.952)--(4.338,6.075)%
  --(4.396,5.979)--(4.455,5.904)--(4.513,5.535)--(4.571,5.567)--(4.630,5.718)--(4.688,5.831)%
  --(4.746,5.753)--(4.805,5.627)--(4.863,5.734)--(4.980,5.640)--(5.038,5.830)--(5.097,5.706)%
  --(5.155,5.611)--(5.213,5.606)--(5.272,5.768)--(5.330,5.801)--(5.388,5.809)--(5.447,5.502)%
  --(5.505,5.520)--(5.563,5.483)--(5.622,5.853)--(5.680,5.857)--(5.738,5.620)--(5.797,5.875)%
  --(5.855,5.859)--(5.913,5.443)--(5.972,5.420)--(6.030,5.728)--(6.088,5.460)--(6.147,5.589)%
  --(6.205,5.599)--(6.264,5.390)--(6.322,5.694)--(6.380,5.636)--(6.439,5.810)--(6.497,5.456)%
  --(6.555,5.755)--(6.614,5.651)--(6.730,5.998)--(6.789,5.367)--(6.847,5.273);
\gpcolor{color=gp lt color border}
\node[gp node right,font={\fontsize{8pt}{9.6pt}\selectfont}] at (5.379,2.551) {en-et after 100k of en-fi};
\gpcolor{rgb color={1.000,0.000,0.000}}
\gpsetlinetype{gp lt plot 2}
\draw[gp path] (5.563,2.551)--(6.479,2.551);
\draw[gp path] (1.634,0.985)--(1.654,5.287)--(1.712,6.059)--(1.771,6.518)--(1.829,6.628)%
  --(1.887,6.884)--(1.946,6.796)--(2.004,6.947)--(2.062,7.138)--(2.121,7.054)--(2.179,7.152)%
  --(2.237,6.917)--(2.296,7.104)--(2.354,7.061)--(2.412,6.886)--(2.471,6.979)--(2.529,6.781)%
  --(2.587,6.607)--(2.646,6.565)--(2.704,6.905)--(2.763,6.586)--(2.821,6.946)--(2.879,6.677)%
  --(2.938,6.681)--(2.996,6.811)--(3.054,6.296)--(3.113,6.658)--(3.171,6.603)--(3.229,6.362)%
  --(3.288,6.538)--(3.346,6.481)--(3.404,6.968)--(3.463,6.391)--(3.521,6.459)--(3.579,6.474)%
  --(3.638,6.434)--(3.696,6.635)--(3.754,6.251)--(3.813,6.463)--(3.871,6.411)--(3.930,6.266)%
  --(3.988,6.561)--(4.046,5.904)--(4.105,6.246)--(4.163,6.291)--(4.221,6.605)--(4.280,6.308)%
  --(4.338,6.070)--(4.396,6.200)--(4.455,6.374)--(4.513,6.144)--(4.571,6.203)--(4.630,6.146)%
  --(4.688,6.083)--(4.746,6.189)--(4.805,6.055)--(4.863,5.965)--(4.921,6.160)--(4.980,6.129)%
  --(5.038,6.025)--(5.097,6.311)--(5.155,6.164)--(5.213,5.953)--(5.272,6.231)--(5.330,6.198)%
  --(5.388,5.813)--(5.447,6.058)--(5.505,5.917)--(5.563,6.254)--(5.622,6.281)--(5.680,5.895)%
  --(5.738,5.840)--(5.797,6.189)--(5.855,6.501)--(5.913,6.202)--(5.972,5.974)--(6.030,5.679)%
  --(6.088,5.867)--(6.147,5.907)--(6.205,5.929)--(6.264,6.077)--(6.322,5.734)--(6.380,6.158)%
  --(6.439,6.052)--(6.497,5.770)--(6.555,5.866)--(6.614,6.095)--(6.672,6.112)--(6.730,6.167)%
  --(6.789,6.098)--(6.847,6.088);
\gpcolor{color=gp lt color border}
\node[gp node right,font={\fontsize{8pt}{9.6pt}\selectfont}] at (5.379,2.243) {en-et after 200k of en-fi};
\gpcolor{rgb color={0.000,0.000,1.000}}
\gpsetlinetype{gp lt plot 3}
\draw[gp path] (5.563,2.243)--(6.479,2.243);
\draw[gp path] (2.215,0.985)--(2.237,5.950)--(2.296,6.689)--(2.354,7.223)--(2.412,7.249)%
  --(2.471,7.530)--(2.529,7.706)--(2.587,7.624)--(2.646,7.670)--(2.704,7.745)--(2.763,7.429)%
  --(2.821,7.437)--(2.879,7.638)--(2.938,7.634)--(2.996,7.358)--(3.054,7.500)--(3.113,7.381)%
  --(3.171,7.296)--(3.229,7.064)--(3.288,7.438)--(3.346,7.331)--(3.404,7.348)--(3.463,7.447)%
  --(3.521,7.663)--(3.579,7.572)--(3.638,7.032)--(3.696,7.045)--(3.754,7.174)--(3.813,7.323)%
  --(3.871,7.344)--(3.930,7.075)--(3.988,7.322)--(4.046,6.736)--(4.105,6.922)--(4.163,7.130)%
  --(4.221,7.142)--(4.280,6.990)--(4.338,7.078)--(4.396,7.284)--(4.455,7.103)--(4.513,6.946)%
  --(4.571,7.017)--(4.630,6.910)--(4.688,6.759)--(4.746,6.633)--(4.805,6.650)--(4.863,6.774)%
  --(4.921,6.609)--(4.980,6.860)--(5.038,6.674)--(5.097,6.966)--(5.155,6.691)--(5.213,6.688)%
  --(5.272,6.404)--(5.330,6.146)--(5.388,6.628)--(5.447,6.694)--(5.505,6.554)--(5.563,6.475)%
  --(5.622,6.256)--(5.680,6.317)--(5.738,6.345)--(5.797,6.587)--(5.855,6.539)--(5.913,6.638)%
  --(5.972,6.469)--(6.205,6.522)--(6.264,6.360)--(6.322,6.575)--(6.380,6.592)--(6.439,6.503)%
  --(6.497,6.552)--(6.555,6.532)--(6.614,6.580)--(6.672,6.291)--(6.730,6.673)--(6.789,6.422)%
  --(6.847,6.371);
\gpcolor{color=gp lt color border}
\node[gp node right,font={\fontsize{8pt}{9.6pt}\selectfont}] at (5.379,1.935) {en-et after 400k of en-fi};
\gpcolor{rgb color={0.753,0.502,1.000}}
\gpsetlinetype{gp lt plot 4}
\draw[gp path] (5.563,1.935)--(6.479,1.935);
\draw[gp path] (3.383,0.985)--(3.404,5.788)--(3.463,6.963)--(3.521,7.426)--(3.579,7.564)%
  --(3.638,8.450)--(3.696,8.137)--(3.754,8.158)--(3.813,8.007)--(3.871,7.890)--(3.930,8.009)%
  --(3.988,7.928)--(4.046,7.818)--(4.105,7.890)--(4.163,7.840)--(4.221,8.195)--(4.280,7.568)%
  --(4.338,7.745)--(4.396,7.963)--(4.455,7.972)--(4.513,7.734)--(4.571,7.915)--(4.630,7.870)%
  --(4.688,7.894)--(4.746,7.722)--(4.805,7.453)--(4.863,7.522)--(4.921,7.511)--(4.980,7.571)%
  --(5.038,7.935)--(5.097,7.803)--(5.155,7.537)--(5.213,7.414)--(5.272,7.555)--(5.330,7.614)%
  --(5.388,7.367)--(5.447,7.512)--(5.505,7.730)--(5.563,7.314)--(5.622,7.145)--(5.680,6.999)%
  --(5.738,7.092)--(5.797,7.203)--(5.855,7.258)--(5.913,6.917)--(5.972,7.047)--(6.030,7.082)%
  --(6.088,6.952)--(6.147,7.197)--(6.205,6.934)--(6.264,6.911)--(6.322,7.171)--(6.380,7.120)%
  --(6.439,7.148)--(6.497,6.942)--(6.555,7.013)--(6.614,6.888)--(6.672,7.021)--(6.730,7.207)%
  --(6.789,6.720)--(6.847,7.116);
\gpcolor{color=gp lt color border}
\node[gp node right,font={\fontsize{8pt}{9.6pt}\selectfont}] at (5.379,1.627) {en-et after 800k of en-fi};
\gpcolor{rgb color={0.647,0.165,0.165}}
\gpsetlinetype{gp lt plot 5}
\draw[gp path] (5.563,1.627)--(6.479,1.627);
\draw[gp path] (5.718,0.985)--(5.738,5.603)--(5.797,6.742)--(5.855,7.370)--(5.913,7.729)%
  --(5.972,8.081)--(6.030,8.019)--(6.088,8.280)--(6.147,8.036)--(6.205,8.216)--(6.264,8.423)%
  --(6.322,8.398)--(6.380,8.183)--(6.439,8.126)--(6.497,8.223)--(6.555,8.356)--(6.614,8.457)%
  --(6.672,7.864)--(6.730,8.155)--(6.789,8.138)--(6.847,8.105);
\gpcolor{color=gp lt color border}
\node[gp node right,font={\fontsize{8pt}{9.6pt}\selectfont}] at (5.379,1.319) {Non-comparable English-Finnish};
\gpcolor{rgb color={0.745,0.745,0.745}}
\gpsetlinetype{gp lt plot 6}
\draw[gp path] (5.563,1.319)--(6.479,1.319);
\draw[gp path] (1.157,0.985)--(1.158,1.060)--(1.304,4.112)--(1.450,5.363)--(1.596,5.619)%
  --(1.741,6.334)--(1.887,6.944)--(2.033,7.485)--(2.179,6.875)--(2.325,7.337)--(2.471,7.342)%
  --(2.617,7.619)--(2.763,7.850)--(2.908,7.840)--(3.054,7.570)--(3.200,7.697)--(3.346,7.710)%
  --(3.492,7.855)--(3.638,7.825)--(3.784,7.796)--(3.930,8.151)--(4.075,8.413)--(4.221,8.261)%
  --(4.367,8.270)--(4.513,7.939)--(4.659,7.983)--(4.805,8.125)--(4.951,8.268)--(5.097,8.152)%
  --(5.242,8.187)--(5.388,8.131)--(5.534,8.354)--(5.680,8.405)--(5.826,8.322)--(5.972,8.553)%
  --(6.118,8.550)--(6.264,8.072)--(6.409,8.534)--(6.555,8.237)--(6.701,8.305)--(6.847,8.599);
\gpcolor{color=gp lt color border}
\gpsetlinetype{gp lt border}
\gpsetlinewidth{1.00}
\draw[gp path] (1.012,8.630)--(1.012,0.985)--(6.847,0.985)--(6.847,8.630)--cycle;
\gpdefrectangularnode{gp plot 1}{\pgfpoint{1.012cm}{0.985cm}}{\pgfpoint{6.847cm}{8.630cm}}
\end{tikzpicture}